\pdfoutput=1

\documentclass[11pt]{article}

\usepackage[preprint]{acl} 

\usepackage{times}
\usepackage{latexsym}

\usepackage[T1]{fontenc}

\usepackage[utf8]{inputenc}

\usepackage{microtype}

\usepackage{inconsolata}

\usepackage{graphicx}

\usepackage{amsmath}        
\usepackage{amssymb}        
\usepackage{algorithm}
\usepackage{algorithmic}
\usepackage{array}          
\usepackage{booktabs}       
\usepackage{amsfonts}       
\usepackage{nicefrac}       
\usepackage{caption}        
\usepackage{subcaption}

\usepackage{url}

\newcommand{\codeurl}{\url{https://github.com/ReLink-Inc/PropRAG}}

\usepackage{tcolorbox}
\tcbuselibrary{listings} 
\tcbuselibrary{breakable} 

\lstdefinestyle{mypromptstyle}{
  basicstyle=\ttfamily\small, 
  breaklines=true,             
  postbreak=\mbox{\textcolor{red}{$\hookrightarrow$}\space}, 
  showstringspaces=false       
}

\newcommand{\code}[1]{\texttt{#1}}

\title{PropRAG: Guiding Retrieval with Beam Search over Proposition Paths}

\author{
  \textbf{Jingjin Wang} \hspace{.3in}
  \textbf{Jiawei Han}\\
  Siebel School of Computing and Data Science, University of Illinois Urbana-Champaign \\
  \texttt{\{jingjin9, hanj\} @illinois.edu} \\}

\begin{document}
\maketitle
\begin{abstract}
Retrieval Augmented Generation (RAG) has become the standard approach for equipping Large Language Models (LLMs) with up-to-date knowledge. However, standard RAG, relying on independent passage retrieval, often fails to capture the interconnected nature of information required for complex, multi-hop reasoning. While structured RAG methods attempt to address this using knowledge graphs built from triples, we argue that the inherent context loss of triples (context collapse) limits the fidelity of the knowledge representation. We introduce \textbf{PropRAG}, a novel RAG framework that shifts from triples to context-rich \textbf{propositions} and introduces an efficient, \textbf{LLM-free online beam search} over proposition paths to discover multi-step reasoning chains. By coupling a higher-fidelity knowledge representation with explicit path discovery, PropRAG achieves state-of-the-art zero-shot Recall@5 and F1 scores on 2Wiki, HotpotQA, and MuSiQue, advancing non-parametric knowledge integration by improving evidence retrieval through richer representation and efficient reasoning path discovery.\footnote{Code and data: \codeurl}
\end{abstract}

\section{Introduction}
\label{sec:introduction}

Large Language Models (LLMs) often struggle with knowledge-intensive tasks, particularly when information is newly emergent or requires complex reasoning over multiple facts. Retrieval Augmented Generation (RAG) \citep{lewis2020retrieval} addresses this by retrieving external knowledge, offering a non-parametric solution that mitigates issues like catastrophic forgetting in continual learning \citep{cohen2024evaluating, gu2024knowledge}. However, conventional RAG systems \citep{karpukhin2020dense, lee2025nv} typically retrieve evidence passages independently based on query similarity. This approach struggles with multi-step queries that require synthesizing interconnected information—a process crucial for sense-making \citep{klein2006making} and associativity \citep{suzuki2007making}.

To improve multi-hop retrieval, structured RAG methods have emerged. Approaches like HippoRAG 2 \citep{gutierrez2025rag} utilize knowledge graphs (KGs) built from (Subject, Predicate, Object) triples and use graph algorithms (e.g., Personalized PageRank) to rank relevant nodes. While effective, we argue these methods suffer from "context collapse." Triples are a lossy compression of natural language, discarding crucial nuances like conditionality, provenance, and n-ary relationships (Section \ref{ssec:prop_extraction_arg}). This loss of fidelity limits the system's ability to reconstruct complex reasoning chains. Other advanced RAG strategies rely on online LLM calls during the retrieval process (e.g., generating subsequent queries) \citep{trivedi2022interleaving, resp2024}, which introduces significant latency, cost, and potential consistency issues.

Our main insight is that effective multi-hop RAG requires two key shifts: (1) moving from lossy triples to a higher-fidelity knowledge representation that preserves context, and (2) moving from simple node ranking to explicit, efficient discovery of reasoning paths.

We introduce \textbf{PropRAG}, a novel RAG framework designed for dynamic, interconnected memory retrieval without requiring online LLM inference during the search process. PropRAG features two core innovations:

\begin{enumerate}
\parskip 0.2ex
    \item \textbf{Propositions as High-Fidelity Knowledge Units:} Extracted offline by an LLM, propositions are atomic, self-contained statements that preserve contextual richness (e.g., conditional clauses, multi-entity events) beyond triples (Section \ref{ssec:prop_extraction_arg}).
    \item \textbf{LLM-Free Online Beam Search for Path Discovery:} A novel two-stage retrieval process culminating in an algorithmic beam search discovers and scores paths of interconnected propositions. This search operates on a pre-built graph using embeddings and graph structure, avoiding the costs and latency of LLM inference during evidence discovery (Section \ref{sec:methodology}).
\end{enumerate}

By coupling context-rich propositions with an efficient, algorithmic path-finding mechanism, PropRAG enhances the system's ability to perform associative reasoning. Experiments demonstrate substantial improvements, particularly on multi-hop QA tasks. PropRAG sets new state-of-the-art zero-shot RAG scores on several challenging benchmarks, advancing the capabilities of non-parametric knowledge integration in LLMs.

\begin{figure*}[t!]
\centering
\includegraphics[width=\textwidth]{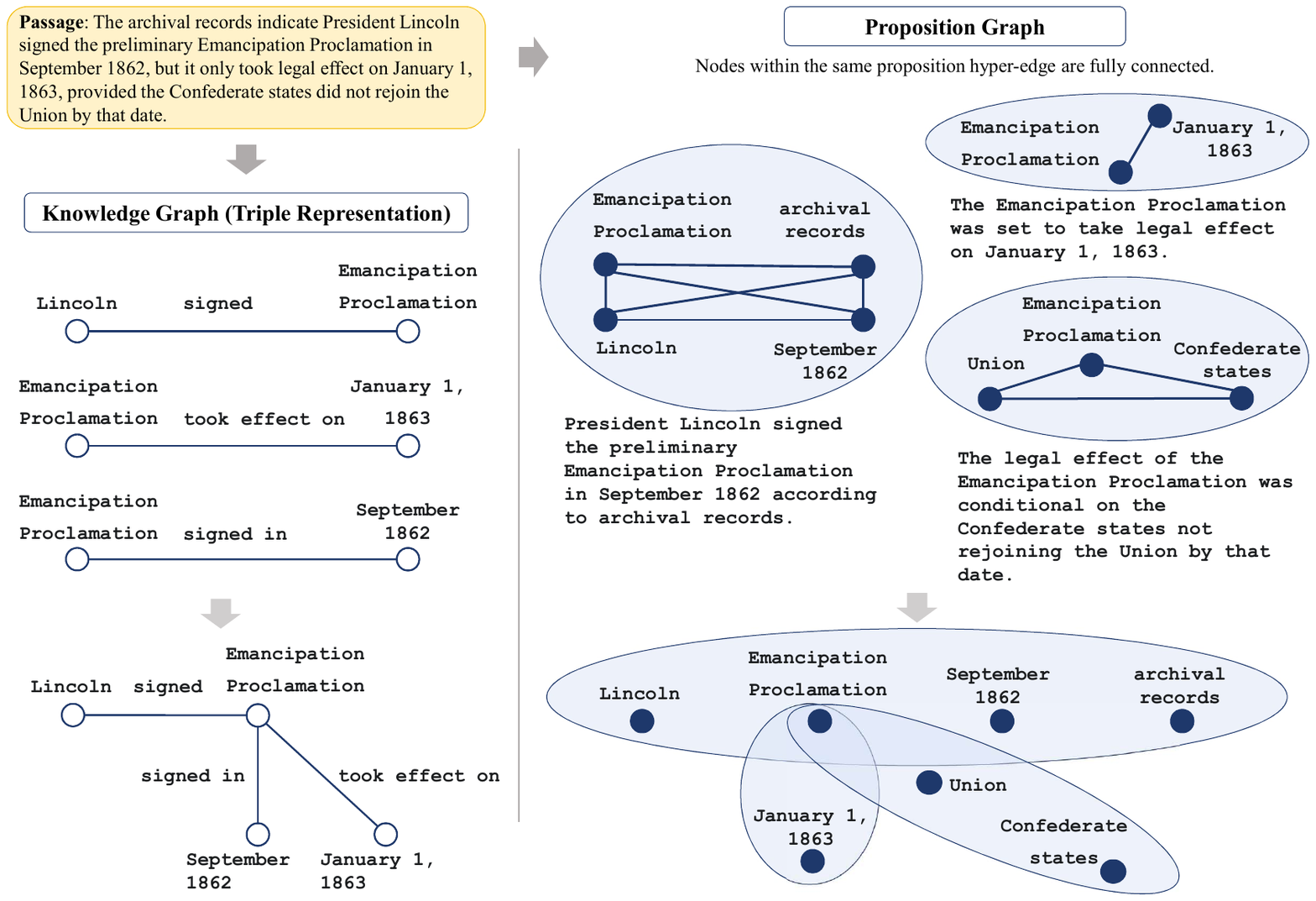}
\caption{Comparison of a traditional Knowledge Graph (KG) versus a Proposition Graph for a complex passage. \textbf{Left:} The triple-based KG struggles to natively represent provenance ("archival records") and conditional clauses. It results in disconnected facts where the crucial context (the conditionality of the Emancipation Proclamation taking effect) is omitted, leading to context loss. \textbf{Right:} The PropRAG proposition graph utilizes implicit hyper-edges (fully connected cliques within shaded ovals) to link all entities co-occurring within a single, context-rich proposition. This structure directly preserves nuances like conditionality and provenance.}
\label{fig:prop_vs_triple}
\end{figure*}

\section{Related Work}
\label{sec:related_work}
\textbf{Retrieval Augmented Generation (RAG)} frameworks \citep{lewis2020retrieval} augment LLMs by retrieving documents. Early methods like DPR \citep{karpukhin2020dense} used embedding similarity. Despite better embeddings \citep{izacard2022unsupervised, ni2022large, lee2025nv}, standard RAG struggles with multi-document synthesis \citep{asai2020learning}.

\textbf{Multi-Hop RAG} aims to address this. Iterative methods \citep{asai2020learning, trivedi2022interleaving} retrieve sequentially, often using LLMs online. Graph-based RAG \citep{edge2024local, sarthi2024raptor} structures knowledge, often using KGs. HippoRAG 2 \citep{gutierrez2025rag} used Personalized PageRank over triple-based KGs, focusing on node ranking. PropRAG differs fundamentally by using context-rich propositions and employing an explicit, LLM-free online beam search focused on discovering complete reasoning paths, rather than just ranking nodes.

\textbf{Beam Search in Retrieval.} 
Recent work has explored leveraging beam search to improve multi-hop retrieval. For instance, \cite{zhang2023end} proposed Beam Retrieval, an end-to-end trainable framework where beam search is used during both training and inference to find optimal passage sequences. Their approach learns a scoring function optimized across hops using ground-truth passage chains.

In contrast, PropRAG adopts a zero-shot online retrieval strategy. While also employing beam search, PropRAG operates over a graph of contextually rich propositions, extracted offline. Our beam search algorithmically discovers proposition paths based on pre-computed embedding similarity and graph connectivity, crucially avoiding online LLM inference costs or task-specific training during the retrieval phase. PropRAG thus decouples complex reasoning path discovery from end-to-end training dependencies, focusing on leveraging richer knowledge units (propositions) and algorithmic path exploration.

\textbf{Propositions as Retrieval Units.} \citet{chen2024dense} demonstrated that retrieving propositions (atomic factoids) can outperform passage or sentence retrieval in standard RAG. PropRAG builds upon this insight by integrating propositions into a graph structure and utilizing beam search to chain them together for multi-hop reasoning.

\section{Propositions: Escaping the Tyranny of the Triple}
\label{ssec:prop_extraction_arg}
The efficacy of multi-hop reasoning is constrained by the fidelity of the underlying knowledge representation. Traditional KG-based RAG systems rely on $\langle \text{Subject, Predicate, Object} \rangle$ triples. We contend this is a lossy compression that discards crucial nuances, leading to "context collapse." Propositions—atomic, self-contained statements that preserve context—offer a higher-fidelity alternative. Figure \ref{fig:prop_vs_triple} illustrates how a proposition graph retains context lost in a standard KG.

Key limitations of triples addressed by propositions include:

\vspace{-.5ex}
\begin{enumerate}
\parskip 0.2ex
\item \textbf{Omission of Conditional Context:} Facts are often contingent upon specific conditions (temporal, spatial, etc.), which triples frequently discard.
\begin{itemize}
\parskip 0.2ex
\item \textit{Example Passage:} The experimental drug showed promise in Phase 2 trials, reducing tumor size significantly, but only in patients under 50 with the specific KRAS mutation.
\item \textit{Triples might yield:} \code{(drug, showed, promise)}, \code{(drug, reduced, tumor size)}. The critical conditions are lost.
\item \textit{Propositions preserve conditions:} E.g., \code{"The significant tumor size reduction by the experimental drug was observed only in patients under 50."}
\end{itemize}

\item \textbf{Limitations in Representing Higher-Order Relations:} Representing complex constructs like provenance, causality, or nested clauses is unnatural in S-P-O format.
    \begin{itemize}
        \parskip 0.2ex
        \item \textit{Triple Limitation:} As shown in Figure \ref{fig:prop_vs_triple} (Left), the triple representation fails to capture the provenance ("Archival records indicate...") and completely omits the critical condition ("provided the Confederate states...") regarding the Emancipation Proclamation. While KGs can use \textit{reification}, this increases complexity and sacrifices clarity.
        \item \textit{Proposition Advantage:} Propositions naturally encapsulate these relations. Figure \ref{fig:prop_vs_triple} (Right) shows these preserved in distinct propositions, modeled using implicit hyper-edges that maintain the full context.
    \end{itemize}

    \item \textbf{Forced Binarization of Unary Predicates (Attributes):} The S-P-O model is fundamentally relational and struggles to represent intrinsic properties (e.g., "The manuscript is fragile").
\begin{itemize}
\parskip 0.2ex
\item \textit{Triple Limitation:} Forcing this into \code{(manuscript, is, fragile)} artificially treats the attribute ("fragile") as an independent object, relying on weak predicates.
\item \textit{Proposition Advantage:} Propositions handle attributes natively: \code{"The ancient manuscript was fragile."}
\end{itemize}

    \item \textbf{Fragmentation of N-ary Relationships (Events):} Real-world events often involve multiple participants. The S-P-O structure forces decomposition into multiple binary triples, fragmenting the event's holistic semantics.
    \begin{itemize}
        \parskip 0.2ex
        \item \textit{Example Passage (Collaboration):} "The groundbreaking research paper on quantum entanglement was co-authored by Alice, Bob, and Charlie in 2023."
        \item \textit{Triples fragment this:} \code{(Alice, co-authored, paper)}, \code{(Bob, co-authored, paper)}, etc., losing that they collaborated *jointly*.
        \item \textit{Proposition Advantage:} A single proposition captures the N-ary relation: \code{"Alice, Bob, and Charlie co-authored the groundbreaking research paper..."} This is represented as an implicit hyper-edge in the proposition graph.
    \end{itemize}
\end{enumerate}

To operationalize this, PropRAG leverages the NLU capabilities of LLMs (Llama-3.3-70B-Instruct) during the \textbf{offline} indexing phase to extract these high-fidelity propositions. This ensures the subsequent online retrieval operates on a contextually rich foundation without further LLM inference for knowledge representation.

\section{Problem Formulation: Finding the Reasoning Path}
\label{sec:problem_formulation}
Traditional RAG aims to retrieve documents $D_{ret}$ maximizing individual relevance $\text{sim}(\text{emb}(d), \text{emb}(q))$. Multi-hop KG-RAG methods like HippoRAG 2 \citep{gutierrez2025rag} typically rank nodes (entities/passages) based on proximity to query seeds or graph centrality (e.g., via PPR), but do not explicitly construct or evaluate multi-step reasoning paths. Furthermore, when KGs use context-poor triples, the semantic richness available for ranking is limited.

We reformulate multi-hop retrieval as finding an optimal \textit{path} of interconnected propositions $P = (p_1, p_2, ..., p_k)$ that collectively answer query $q$. Propositions $p_i, p_{i+1}$ are linked if they share common or synonymous entities. This connection occurs naturally in the proposition graph (Figure \ref{fig:prop_vs_triple}, right), where entities shared between proposition hyper-edges act as bridges.

The objective is to find \begin{equation}P^* = \underset{P \in \text{ConnectedPaths}(\mathcal{P})}{\text{argmax}} \text{ Score}(P, q),\end{equation} where $\text{Score}(P, q)$ measures the relevance of the entire path, potentially via aggregated or concatenated embeddings. As finding the global optimum over all possible paths is intractable, we employ beam search as an efficient heuristic.

\begin{figure*}[t!]
\centering
\includegraphics[width=\textwidth]{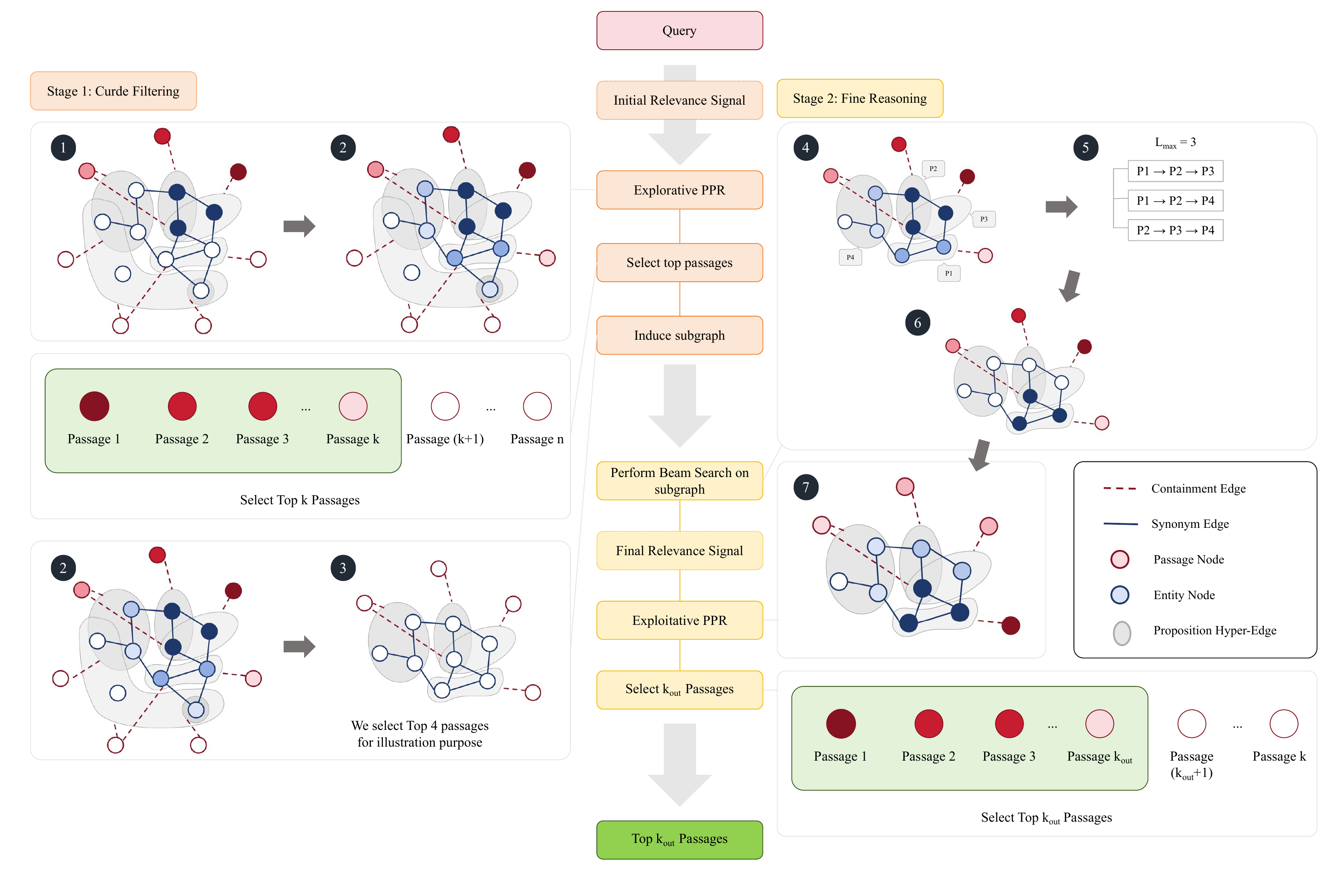}
\caption{The two-stage online retrieval architecture of PropRAG. \textbf{Stage 1 (Coarse Filtering):} Employs exploratory PPR (high damping factor) on the full proposition graph $G$ to induce a focused, relevant subgraph ($G_{sub}$). \textbf{Stage 2 (Fine Reasoning):} Executes a graph-guided beam search on $G_{sub}$ to discover explicit reasoning paths (illustrated in Figure \ref{fig:qualitative_beam_main}), generates refined relevance signals based on these paths, applies exploitative PPR (low damping factor) on $G_{sub}$ using the refined signals, and selects the final top-$k_{out}$ evidence passages.}
\label{fig:propRAG_overview}
\end{figure*}

\begin{figure}[t!]
\centering
\fbox{
\begin{minipage}{\dimexpr\columnwidth-2\fboxsep-2\fboxrule\relax} 
\small 
\textbf{Query (Q):} What year did the Governor of the city where the basilica named after the same saint as the one that Mantua Cathedral is dedicated to die? \\
\textbf{Gold Answer:} 1952

\textbf{Initial Propositions (Stage 1 Retrieval):}
\begin{itemize} \itemsep0pt \parskip0pt \leftmargin=5pt
    \item (P1) 0.4275 - Mantua Cathedral is a Roman Catholic cathedral dedicated to Saint Peter.
    \item 0.3851 - Mantua Cathedral is the seat of the Bishop of Mantua.
    \item ...
    \item (P3) 0.3135 - No successor was appointed to the post of Governor of Vatican City after Marchese Camillo Serafini's death in 1952.
\end{itemize}

\textbf{Beam Search (Stage 2, Depth 2/3):}
\textit{Path expansion from P1 .}
\begin{itemize} \itemsep0pt \parskip0pt \leftmargin=5pt
    \item 0.4792 - (P1) Mantua Cathedral... dedicated to Saint Peter. $\rightarrow$ (P2) St. Peter's Basilica is located in Vatican City. (via entity link: "Saint Peter" $\rightarrow$ "St. Peter")
    \item 0.4756 - (P1) $\rightarrow$ Alfredo Ormando died on 23 January 1998 in Rome. (via entity link: "Roman" $\rightarrow$ "Rome")
    \item ...
\end{itemize}

\textbf{Beam Search (Stage 2, Depth 3/3):}
\textit{Path expansion from P2.}
\begin{itemize} \itemsep0pt \parskip0pt \leftmargin=5pt
    \item \textbf{0.5989 - (P1) Mantua Cathedral... $\rightarrow$ (P2) St. Peter's Basilica is located in Vatican City. $\rightarrow$ (P3) No successor was appointed... after Marchese Camillo Serafini's death in 1952.} (via entity link: "Vatican City")
    \item 0.5674 - (P1) $\rightarrow$ (P2) $\rightarrow$ Marchese Camillo Serafini held the post of Governor... until his death in 1952.
    \item ...
\end{itemize}
\textbf{Observation:} The crucial intermediate proposition P2 exhibited low initial relevance to Q but was discovered via the graph-guided beam search, enabling the construction of the complete reasoning path P1 $\rightarrow$ P2 $\rightarrow$ P3.
\end{minipage}
}
\caption{Running Example: Beam search execution ($L_{max}=3$) for a MuSiQue query, illustrating the discovery of a multi-hop reasoning path in PropRAG. Proposition text is abridged for clarity.}
\label{fig:qualitative_beam_main}
\end{figure}

\section{Methodology: PropRAG}
\label{sec:methodology}
The PropRAG framework operationalizes the path-finding objective (Section \ref{sec:problem_formulation}) through a structured offline indexing phase followed by an efficient, two-stage online retrieval process (Figure \ref{fig:propRAG_overview}). We utilize the complex multi-hop query presented in Figure \ref{fig:qualitative_beam_main} as a running example to illustrate the methodology.

\subsection{Offline Indexing (LLM-assisted)}
The objective of the offline indexing phase is to construct the high-fidelity proposition graph, a prerequisite for efficient online path discovery.
\vspace{-.5ex}
\begin{enumerate}
\parskip 0.2ex
    \item \textbf{Knowledge Extraction:} An LLM (e.g., Llama-3.3-70B-Instruct) is employed to extract propositions and their constituent entities from the corpus $\mathcal{D}$ (Prompts detailed in Appendix \ref{sec:appendix_prompts_app}).
    \item \textbf{Graph Construction:} A proposition graph $G=(V,E)$ is constructed. Nodes $V$ represent entities and passages. Edges $E$ link entities if they co-occur within the same proposition (forming implicit hyper-edges) or if they are determined to be synonymous via embedding similarity (Details in Appendix \ref{app:graph_construction}).
    \item \textbf{Embedding Generation:} Embeddings (e.g., using NV-Embed-v2) are computed and stored for all nodes and propositions.
\end{enumerate}

\subsection{Online Retrieval (LLM-free)}
The online retrieval component implements a two-stage, LLM-free architecture designed to identify salient reasoning paths. This coarse-to-fine optimization strategy balances the need for broad exploration of the graph with focused refinement of candidate paths.

\subsubsection{Stage 1: Coarse-grained Subgraph Induction}
The objective of Stage 1 is to efficiently prune the search space by inducing a highly relevant, localized subgraph $G_{sub}$ from the global graph $G$.

\begin{enumerate}
    \item \textbf{Initial Candidate Retrieval:} The top-$N_{prop}$ propositions most semantically similar to the query $q$ are retrieved.
    \textit{(Running Example, Fig. \ref{fig:qualitative_beam_main})}: Propositions P1 (Mantua Cathedral) and P3 (Governor's death) are retrieved due to high initial similarity, while the necessary intermediate P2 is not highly ranked.
    \item \textbf{Seed Set Initialization ($S_{initial}$):} Entities $\mathcal{E}(p)$ contained within these top propositions are extracted. The top-$N_{entity}$ unique entities form the initial seed set $S_{initial}$. Uniform weights are assigned to promote diverse graph exploration.
    \item \textbf{Exploratory Graph Traversal (PPR):} Personalized PageRank (PPR) is executed on the global graph $G$, with the reset probability distribution concentrated on $S_{initial}$. A high damping factor (e.g., 0.75) is utilized to facilitate broader traversal originating from these seeds.
    \item \textbf{Subgraph Induction ($G_{sub}$):} The top-$K$ passages identified by the PPR scores, along with their associated entities and propositions in $G$, constitute the focused subgraph $G_{sub}$. This localized graph serves as the operational search space for Stage 2.
\end{enumerate}

\subsubsection{Stage 2: Fine-grained Path Discovery and Ranking}
The objective of Stage 2 is to explicitly discover multi-proposition reasoning paths within $G_{sub}$ and leverage these paths to determine the final ranking of evidence passages.

\begin{enumerate}
    \item \textbf{Beam Search Path Exploration:} The beam search algorithm (detailed in Section \ref{sssec:beam_search}) systematically explores $G_{sub}$ to identify connected proposition paths up to a maximum length $L_{max}$. It maintains a beam of the top-$B$ paths ranked by relevance to the query $q$.
    \textit{(Running Example, Fig. \ref{fig:qualitative_beam_main})}: Initiating from P1, the beam search traverses the graph via the "Saint Peter" entity link, discovering P2 ("St. Peter's Basilica is located in Vatican City"). Subsequently, it connects P2 to P3 via "Vatican City", constructing the complete path P1 $\rightarrow$ P2 $\rightarrow$ P3.
    \item \textbf{Refined Seed Set Construction ($S_{final}$):} A refined seed set is generated by integrating relevance signals from two sources: \textit{Exploration Seeds} (derived from initial query similarity) and \textit{Exploitation Seeds} (entities central to the top-scoring paths identified by the beam search).
    \item \textbf{Exploitative Path-informed Ranking (PPR):} A second iteration of PPR is executed, constrained to $G_{sub}$. The reset probability vector incorporates normalized scores from both exploration and exploitation seeds (Appendix \ref{app:entity_score_calculation}). A lower damping factor (e.g., 0.45) is employed to intensify the focus on these refined relevance signals.
    \item \textbf{Final Evidence Ranking:} Passages are ranked according to their final PPR scores, prioritizing those strongly connected to the salient reasoning paths discovered during the beam search.
\end{enumerate}

\subsection{Beam Search Algorithm for Path Discovery}
\label{sssec:beam_search}
The core of Stage 2 is a specialized beam search algorithm designed to efficiently explore the proposition graph and identify high-relevance reasoning chains without requiring online LLM inference. It heuristically seeks paths $P = (p_1, ..., p_L)$ that maximize the relevance score $\text{Score}(P, q)$.

\textbf{Algorithmic Procedure:}
\begin{itemize}
    \item \textbf{Initialization:} The beam is initialized with paths of length 1, comprising the propositions most semantically similar to the query $q$.
    \item \textbf{Path Expansion (Graph-Guided):} At each step, paths are extended by identifying candidate next propositions $p_{next}$. Candidates are primarily selected from propositions connected to the current path's terminal proposition $p_k$ via shared or synonymous entity links within $G_{sub}$.
    \textit{(Running Example, Fig. \ref{fig:qualitative_beam_main})}: P1 is expanded to P2 due to the shared entity "Saint Peter".
    To mitigate the risk of local optima, the top-3 initial query-relevant propositions are also included as candidates ("jump points"), regardless of direct connectivity to $p_k$.
    \item \textbf{Path Scoring (Online Embedding Computation):} The relevance of an expanded path $P_{new} = (P, p_{next})$ is estimated as $\text{Score}(P_{new}, q) \approx \text{sim}(\text{emb}(P_{new}), \text{emb}(q))$. A computationally efficient average embedding is used for preliminary scoring. The top-$M$ candidates are subsequently re-ranked using a more precise score derived from an embedding of the concatenated text of all propositions in $P_{new}$. This step involves efficient online computation using the embedding model, distinct from LLM inference.
    \item \textbf{Selection and Pruning:} From all expanded candidates, the top-$B$ (beam width) paths with the highest scores are retained for the next iteration; others are pruned.
    \item \textbf{Termination:} The search concludes when paths reach the maximum length $L_{max}$ or no further valid expansions are possible.
\end{itemize}
This online beam search operates exclusively on the pre-computed graph structure and efficient online embedding computations. Crucially, it ensures low-latency inference by \textbf{avoiding any costly online LLM calls} for path generation or scoring.

\begin{table*}[t!]
\centering
\caption{Passage Retrieval Performance (Recall@5). Baselines from \citet{gutierrez2025rag}. Best overall result in bold.}
\label{tab:retrieval_results_updated_main}
\small
\begin{tabular}{lccccc}
\toprule
Method & NQ & PopQA & MuSiQue & 2Wiki & HotpotQA \\
\midrule
\textit{Simple Baselines} &&&&& \\
BM25 & 56.1\% & 35.7\% & 43.5\% & 65.3\% & 74.8\% \\
Contriever & 54.6\% & 43.2\% & 46.6\% & 57.5\% & 75.3\% \\
GTR (T5-base) & 63.4\% & 49.4\% & 49.1\% & 67.9\% & 73.9\% \\
\midrule
\textit{Large Embedding Models (Base Retriever)} &&&&& \\
GTE-Qwen2-7B & 74.3\% & 50.6\% & 63.6\% & 74.8\% & 89.1\% \\
GritLM-7B & 76.6\% & 50.1\% & 65.9\% & 76.0\% & 92.4\% \\
NV-Embed-v2 (7B) & 75.4\% & 51.0\% & 69.7\% & 76.5\% & 94.5\% \\
\midrule
\textit{Structure-Augmented RAG} &&&&& \\
RAPTOR & 68.3\% & 48.7\% & 57.8\% & 66.2\% & 86.9\% \\
HippoRAG & 44.4\% & 53.8\% & 53.2\% & 90.4\% & 77.3\% \\
HippoRAG 2 & 78.0\% & 51.7\% & 74.7\% & 90.4\% & 96.3\% \\
\midrule
\textbf{PropRAG (Ours)} &&&&& \\
\quad $L_{max}=1$ (No beam search) & \textbf{78.4\%} & \textbf{56.3\%} & 75.6\% & 92.0\% & 95.7\% \\
\quad $L_{max}=2$ & 78.1\% & 56.1\% & 77.6\% & 93.4\% & 97.2\% \\
\quad $\boldsymbol{L_{max}=3}$ \textbf{(Default)} & 77.9\% & 56.2\% & \textbf{78.3\%} & \textbf{94.1\%} & \textbf{97.4\%} \\
\quad $L_{max}=4$ & 77.8\% & 56.0\% & 77.6\% & 93.7\% & 97.0\% \\
\bottomrule
\end{tabular}
\end{table*}

\begin{table*}[t!]
\centering
\caption{End-to-End QA Performance (F1 Score) with Llama-3.3-70B-Instruct Reader. Baselines from \citet{gutierrez2025rag}. Best overall result in bold.}
\label{tab:qa_results_updated_main}
\small
\begin{tabular}{lcccccc}
\toprule
Method & NQ & PopQA & MuSiQue & 2Wiki & HotpotQA & Avg \\
\midrule
\textit{No Retrieval (Parametric)} &&&&&& \\
Llama-3.3-70B-Instruct & 54.9\% & 32.5\% & 26.1\% & 42.8\% & 47.3\% & 40.7\% \\
\midrule
\textit{Simple Baselines} &&&&&& \\
Contriever & 58.9\% & 53.1\% & 31.3\% & 41.9\% & 62.3\% & 49.5\% \\
GTR (T5-base) & 59.9\% & 56.2\% & 34.6\% & 52.8\% & 62.8\% & 53.3\% \\
\midrule
\textit{Large Embedding Models (Base Retriever)} &&&&&& \\
GTE-Qwen2-7B & 62.0\% & 56.3\% & 40.9\% & 60.0\% & 71.0\% & 58.0\% \\
GritLM-7B & 61.3\% & 55.8\% & 44.8\% & 60.6\% & 73.3\% & 59.2\% \\
NV-Embed-v2 (7B) & 61.9\% & 55.7\% & 45.7\% & 61.5\% & 75.3\% & 58.0\% \\
\midrule
\textit{Structure-Augmented RAG} &&&&&& \\
RAPTOR & 50.7\% & 56.2\% & 28.9\% & 52.1\% & 69.5\% & 51.5\% \\
GraphRAG & 46.9\% & 48.1\% & 38.5\% & 58.6\% & 68.6\% & 52.1\% \\
LightRAG & 16.6\% & 2.4\% & 1.6\% & 11.6\% & 2.4\% & 6.9\% \\
HippoRAG & 55.3\% & 55.9\% & 35.1\% & 71.8\% & 63.5\% & 56.3\% \\
HippoRAG 2 & \textbf{63.3\%} & 56.2\% & 48.6\% & 71.0\% & 75.5\% & 62.9\% \\
\midrule
\textbf{PropRAG (Ours)} &&&&&& \\
\quad $L_{max}=1$ (No beam search) & 62.2\% & 56.1\% & 52.6\% & 73.5\% & 75.7\% & 64.0\% \\
\quad $L_{max}=2$ & 61.9\% & 56.1\% & 53.4\% & 74.9\% & 76.0\% & 64.4\% \\
\quad $\boldsymbol{L_{max}=3}$ \textbf{(Default)} & 62.5\% & \textbf{56.4\%} & \textbf{53.9\%} & \textbf{75.3\%} & \textbf{76.1\%} & \textbf{64.9\%} \\
\quad $L_{max}=4$ & 62.8\% & 56.0\% & 53.0\% & 75.3\% & 76.1\% & 64.7\% \\
\bottomrule
\end{tabular}
\end{table*}

\begin{table*}[htbp]
\centering
\caption{Ablation Study on Beam Width ($B$) using $L_{max}=3$. Default $B=4$. Performance shown as Recall@5 / F1 Score.}
\label{tab:ablation_beam_width_app_main}
\small
\begin{tabular}{@{}lcccc@{}}
\toprule
Beam Width ($B$) & MuSiQue (R@5 / F1) & 2Wiki (R@5 / F1) & HotpotQA (R@5 / F1) & Average (R@5 / F1) \\
\midrule
1 (Greedy Search) & 76.6\% / 52.9\% & 92.1\% / 74.5\% & 97.0\% / 76.2\% & 88.5\% / 67.9\% \\
2 & 77.4\% / 52.9\% & \textbf{94.4\%} / \textbf{75.8\%} & 97.4\% / 75.7\% & 89.7\% / 68.1\% \\
3 & 78.0\% / 53.1\% & 94.2\% / 75.7\% & \textbf{97.5\%} / \textbf{76.2\%} & 89.9\% / 68.3\% \\
\textbf{4 (Default)} & \textbf{78.3\%} / 53.9\% & 94.1\% / 75.3\% & 97.4\% / 76.1\% & \textbf{89.9\%} / 68.5\% \\
5 & 77.8\% / \textbf{54.4\%} & 93.7\% / 75.4\% & 97.4\% / 76.1\% & 89.6\% / \textbf{68.6\%} \\
6 & 77.8\% / 53.9\% & 93.2\% / 74.6\% & 97.2\% / 75.7\% & 89.4\% / 68.1\% \\
\bottomrule
\end{tabular}
\end{table*}

\begin{table*}[htbp]
\centering
\caption{Full Ablation Study Results (Recall@5). PropRAG uses $L_{max}=3, B=4$ unless noted. HippoRAG 2 (no filter) results from \citet{gutierrez2025rag} Table 4.}
\label{tab:ablation_results_final_app_main}
\small
\begin{tabular}{lcccc}
\toprule
Configuration & MuSiQue & 2Wiki & HotpotQA & Avg\\
\midrule
\textbf{Full PropRAG} & \textbf{78.3\%} & \textbf{94.1\%} & \textbf{97.4\%} & \textbf{89.9\%} \\
\midrule
\textit{Comparison Baselines (Impact of Propositions)} & & & & \\
\quad PropRAG (Propositions only, no two-stage PPR, no beam search) & 75.4\% & 90.4\% & 95.9\% & 87.2\% \\
\quad HippoRAG 2 (Triples, no LLM rerank, comparable setting) & 73.0\% & 90.7\% & 95.4\% & 86.4\% \\
\midrule
\textit{Retrieval Strategy Ablations (Impact of Special Beam Search Settings)} & & & & \\
\quad Exploration Seeds Only & 75.6\% & 92.0\% & 95.6\% & 87.7\%\\
\quad Exploitation Seeds Only & 77.9\% & 91.4\% & 97.6\% & 89.0\%\\
\quad No Graph Guidance (Similarity only) & 77.4\% & 92.9\% & 96.6\% & 89.0\%\\
\bottomrule
\end{tabular}
\end{table*}

\section{Experiments}
\label{sec:experiments}

\subsection{Setup}
\textbf{Datasets:} We evaluate on five diverse QA datasets: NaturalQuestions (NQ) \citep{wang2024rear} and PopQA \citep{mallen2023not} for single-hop reasoning, and 2WikiMultihopQA (2Wiki) \citep{ho20202wikimultihopqa}, HotpotQA \citep{yang2018hotpotqa}, and MuSiQue-Ans \citep{trivedi2022musique} for multi-hop reasoning. We use the 1000-query samples and associated corpora provided by \citet{gutierrez2025rag} for direct comparability and to manage experimental costs.

\textbf{Baselines:} We compare against a comprehensive set of baselines, including classic retrievers (BM25, Contriever, GTR), large embedding models serving as base retrievers (GTE-Qwen2, GritLM, NV-Embed-v2), and leading structure-augmented RAG methods (RAPTOR, GraphRAG, LightRAG, HippoRAG, HippoRAG 2). Baseline results are primarily adopted from \citet{gutierrez2025rag}.

\textbf{Implementation and Reproducibility:} PropRAG uses Llama-3.3-70B-Instruct \citep{llama3modelcard} for offline proposition extraction and as the QA reader, and NV-Embed-v2 (7B) \citep{lee2025nv} for all embeddings. Key parameters include beam width $B=4$ and max path length $L_{max}=3$. A full list of hyperparameters is provided in Appendix \ref{app:imp_details}. 

\textbf{Metrics:} We report standard metrics: Passage Recall@5 for retrieval quality, and QA F1 score for end-to-end performance, following MuSiQue evaluation scripts \citep{trivedi2022musique}.

\subsection{Results and Discussion}
We present the main retrieval (Recall@5) and end-to-end QA (F1 Score) results in Table \ref{tab:retrieval_results_updated_main} and Table \ref{tab:qa_results_updated_main}, respectively.

\textbf{Overall Performance:} PropRAG (Default: $L_{max}=3, B=4$) achieves a state-of-the-art average F1 score of 64.9\%. This outperforms the previous best structured RAG, HippoRAG 2, by 2.0 points, and the base retriever, NV-Embed-v2, by a substantial 6.9 points. This highlights the synergy of context-rich propositions and explicit, LLM-free online path discovery.

\textbf{The Benefit of Propositions ($L_{max}=1$):} Even without multi-step path expansion ($L_{max}=1$, effectively disabling the beam search but retaining the two-stage PPR on the proposition graph), PropRAG significantly outperforms baselines. For instance, on MuSiQue, $L_{max}=1$ achieves 52.6\% F1, a +4.0\% improvement over HippoRAG 2 (48.6\%). This demonstrates the inherent benefit of using the higher-fidelity proposition graph structure, even before extensive beam search exploration.

\textbf{The Benefit of Beam Search ($L_{max} > 1$):} Explicit path discovery via beam search further boosts performance, particularly on multi-hop datasets. Comparing the default configuration ($L_{max}=3$) to the $L_{max}=1$ ablation, we observe gains in average F1 by +0.9\% (to 64.9\%). The impact on retrieval is even more pronounced, with MuSiQue Recall@5 increasing by +2.7\% (from 75.6\% to 78.3\%). This confirms that exploring 2-3 hop paths uncovers crucial evidence often missed by methods focused only on initial similarity. We find $L_{max}=3$ provides the optimal balance.

\textbf{Performance on Simpler Tasks:} On single-hop datasets NQ and PopQA, PropRAG remains robust (e.g., 62.5\% NQ F1, 56.4\% PopQA F1 with $L_{max}=3$), demonstrating that the added complexity does not degrade performance on tasks with less multi-hop dependency.

\subsection{Ablation Studies}
\label{sec:ablation_studies_main}
We conduct ablation studies on the multi-hop datasets (MuSiQue, 2Wiki, HotpotQA) to validate key design choices. Tables \ref{tab:ablation_beam_width_app_main} and \ref{tab:ablation_results_final_app_main} show our findings.

\textbf{Propositions vs. Triples:} To isolate the impact of the knowledge representation, we compare PropRAG using only its first stage PPR (PPR on the proposition graph) against HippoRAG 2 in a comparable setting (no filter, using triples and PPR; results from \citet{gutierrez2025rag} Table 4). PropRAG (Stage 1 PPR only) achieves an average Recall@5 of 87.2\%, compared to 86.4\% for HippoRAG 2 (+0.8\%). The difference is most notable on MuSiQue (75.4\% vs. 73.0\%, +2.4\%). This confirms that the richer context captured in propositions provides a stronger foundation for graph-based retrieval, even without multi-step beam search.

\textbf{Impact of Beam Width ($B$):} We vary the beam width $B$ while keeping $L_{max}=3$. Increasing $B$ from 1 (greedy search) to 4 (default) improves average R@5 by +1.4\% (from 88.5\% to 89.9\%) and average F1 by +0.6\%. $B=4$ offers a strong balance between exploration and computational cost.

\textbf{Importance of Graph Guidance:} We test the importance of the graph structure guiding the beam search by allowing path expansion based purely on embedding similarity, without requiring graph connectivity (beyond the initial "jump points"). This configuration ("No Graph Guidance") achieves an average R@5 of 89.0\%, which is 0.9\% lower than the full PropRAG (89.9\%). The drop is significant on 2Wiki (92.9\% vs. 94.1\%). This demonstrates that leveraging the explicit connections in the proposition graph effectively guides the search towards more coherent reasoning paths, rather than relying solely on semantic similarity which can be noisy.
    
These ablations collectively confirm the contributions of the higher-fidelity proposition representation, the effectiveness of the graph-guided beam search, and the robustness of the two-stage retrieval architecture.

\subsection{Efficiency Analysis}
\label{sec:efficiency_analysis}
A key design goal of PropRAG is to enable sophisticated multi-hop reasoning efficiently by strategically investing in offline knowledge structuring to avoid costly online LLM calls.

\textbf{Offline Cost-Benefit Trade-off:} The offline indexing phase requires LLM calls for proposition extraction. As detailed in Appendix \ref{app:cost_efficiency} (Table \ref{tab:token_cost_comparison_app}), PropRAG required 21.1M total LLM tokens (Input+Output) to index the MuSiQue dataset. While this is higher than HippoRAG 2 (12.2M tokens), it is significantly lower than other structured RAG methods like GraphRAG (151.6M tokens) and LightRAG (86.8M tokens). Crucially, PropRAG achieves vastly superior performance (53.9\% F1 on MuSiQue) compared to GraphRAG (38.5\%) and LightRAG (1.6\%). PropRAG thus achieves a superior cost-benefit balance, creating higher-quality knowledge units with less computational overhead than comparable methods.

\textbf{Online Latency:} While PropRAG avoids online LLM calls, the two-stage PPR and beam search (which includes online embedding computations for path scoring) add latency compared to a standard dense retriever. We quantify this trade-off through a runtime comparison on the MuSiQue dataset using our hardware (NVIDIA RTX 4090). A standard dense retriever (NV-Embed-v2) takes approximately 50ms per query. PropRAG takes approximately 500-1000ms per query. In contrast, an LLM-dependent retrieval step (e.g., using Llama-3-70B to generate a next-hop query) typically takes 2-5 seconds or more per hop. Therefore, PropRAG is significantly faster (2-10x) online than LLM-in-the-loop retrieval alternatives while providing superior retrieval quality.

\section{Conclusion}
\label{sec:conclusion}
PropRAG represents a significant advancement in RAG by addressing the limitations of both standard retrieval and traditional structured RAG. By shifting from context-poor triples to richer propositions, PropRAG creates a high-fidelity knowledge representation. Coupled with a novel, LLM-free online beam search mechanism, it enables the explicit discovery of multi-step reasoning paths. This dual approach demonstrably improves the quality of retrieved evidence, particularly for complex multi-hop queries. Our experiments show that PropRAG sets new state-of-the-art results for zero-shot RAG systems on several challenging benchmarks. PropRAG underscores the value of explicit, algorithmic modeling of reasoning processes over high-fidelity, pre-structured knowledge, offering a promising direction for developing LLMs with more robust and associative non-parametric memory.

\section*{Acknowledgments}
Research was supported in part by the National Science Foundation IIS-19-56151, NSF IIS 25-37827, the Molecule Maker Lab Institute: An AI Research Institutes program supported by NSF under Award No. 2019897, and the Institute for Geospatial Understanding through an Integrative Discovery Environment (I-GUIDE) by NSF under BRIES Program No. HR0011-24-3-0325. The views and conclusions contained in this paper are those of the authors and should not be interpreted as representing any funding agencies. We also wish to thank Manting Liao for creating the figures.

\section*{Limitations}
PropRAG's primary limitations include the computational overhead of beam search, which, while LLM-free online, is more intensive than simpler retrieval methods. The system's performance is sensitive to the quality of the offline proposition extraction phase; errors or omissions here can propagate. Although online LLM calls are avoided during retrieval, the initial proposition generation relies on an LLM, and its quality can influence downstream results. Furthermore, the graph construction process, particularly the accuracy of entity linking and synonymy detection, plays a crucial role and can be a source of error. The current path scoring relies on embedding similarity, which might not capture all semantic nuances required for perfect path evaluation.

\bibliography{references} 

\appendix
\clearpage 

\section{Appendix}

\subsection{Proposition Graph Construction Details}
\label{app:graph_construction}
The PropRAG Proposition graph $G=(V, E)$ is constructed to facilitate reasoning over interconnected propositions. The vertex set $V$ comprises two main types of nodes:
\begin{itemize}
    \item $V_{entity}$: Nodes representing entities extracted from the text corpus.
    \item $V_{passage}$: Nodes representing the original text passages from which propositions and entities were derived.
\end{itemize}
The edge set $E$ includes the following key types, designed to capture relationships within and between propositions, and to link entities back to their source contexts:
\begin{itemize}
    \item \textbf{Entity Clique Edges (Implicit Proposition Hyper-edge):} For each proposition $p$ extracted from the corpus, which contains a set of entities $\mathcal{E}(p)$, we add undirected edges connecting all pairs of distinct entities $\{e_i, e_j\}$ such that $e_i, e_j \in \mathcal{E}(p)$ and $e_i \neq e_j$. This forms a clique (a fully connected subgraph) among all entities co-occurring within that single proposition. This clique structure implicitly represents the proposition $p$ as a hyper-edge, contextually linking all its constituent entities together, rather than relying on potentially ambiguous predicate-labeled edges between only two entities as in traditional triple stores.
    \item \textbf{Passage Containment Edges:} An undirected edge connects each entity node $e \in V_{entity}$ to the passage node $d \in V_{passage}$ corresponding to the text passage from which entity $e$ (and its associated propositions) were originally extracted. These edges ground entities and propositions in their source documents.
    \item \textbf{Synonymy Edges:} An undirected edge connects two distinct entity nodes $e_i, e_j \in V_{entity}$ if their pre-computed embeddings are highly similar, i.e., $\text{sim}(\text{emb}(e_i), \text{emb}(e_j)) \ge \tau_{syn}$, where $\tau_{syn}$ is a predefined similarity threshold. These edges help bridge different textual mentions of the same underlying concept.
\end{itemize}
This graph structure allows for traversal algorithms (like PPR and beam search) to navigate through the rich context embedded in propositions (via the entity cliques/hyper-edges) and to connect entities back to their original passages, facilitating comprehensive evidence aggregation.

\subsection{Implementation Details}
\label{app:imp_details}
PropRAG leverages Llama-3.3-70B-Instruct for offline proposition extraction (and as the final QA reader for experiments) and NV-Embed-v2 (7B) as the base embedding model for passages, entities, and propositions, ensuring consistency with the HippoRAG 2 baseline setup.
Default parameters used in PropRAG experiments are as follows:
\begin{itemize}
\parskip 0.2ex
    \item Beam width for path discovery ($B$): 4
    \item Maximum path length for beam search ($L_{max}$): 3
    \item Initial PPR damping factor (Stage 1, exploration): 0.75
    \item Final PPR damping factor (Stage 2, exploitation): 0.45
    \item Number of passages in subgraph ($K$): 50
    \item Number of top paths for exact scoring (beam search internal re-ranking) ($M$): 40
    \item Number of top initial seeds for final PPR ($B_{initial}$): 5
    \item Number of top propositions to select seeds from for final PPR ($P_{initial}$): $B$ (Beam width)
    \item Number of top beam-derived seeds for final PPR ($B_{beam}$): 5
    \item Number of top beam-derived paths to select seeds from for final PPR ($P_{beam}$): 5
    \item Synonymy embedding similarity threshold ($\tau_{syn}$): 0.8
    \item Number of initial propositions for seeding Stage 1 PPR ($N_{prop}$): 20
    \item Number of initial entities from the top-$N_{prop}$ propositions for seeding Stage 1 PPR ($N_{entity}$): 40
    \item Weight for passage direct retrieval score in final PPR ($\lambda_{passage}$): 0.05
\end{itemize}
These parameters were determined based on empirical performance on development sets or adopted from common practices in related research where applicable. The choice of $L_{max}=3$ was based on achieving the best average F1 score across development datasets.

\subsection{LLM Prompts}
\label{sec:appendix_prompts_app}

This section details the prompts used for entity and proposition extraction with Llama-3.3-70B-Instruct, crucial for the offline indexing phase of PropRAG.

\begin{figure*}[htbp] 
\begin{tcolorbox}[colback=white, colframe=black, boxrule=0.5pt, sharp corners, left=2mm, right=2mm, top=2mm, bottom=2mm, title=Entity Extraction Prompt] 
\textbf{Instruction:}
Your task is to extract entities from the given paragraph.
Respond with a JSON dictionary only, with a "entities" key that maps to an non-empty list of entities.
All named entities and dates must be included in the list.
All generic entities important to the theme of the passage must be included in the list.
All entities that is involved in a predicate relation to the above entities must be included in the list.
All dates must be included in the list.

\medskip
\hrulefill \par
\textbf{Demonstration:} \par
\textit{Example Paragraph:}
Radio City
Radio City is India's first private FM radio station and was started on 3 July 2001.
It plays Hindi, English and regional songs.
Radio City recently forayed into New Media in May 2008 with the launch of a music portal
- PlanetRadiocity.com that offers music related news, videos, songs, and other
music-related features.

\textit{Example Output:}
\begin{lstlisting}[style=mypromptstyle, framerule=0pt] 
{"entities":
    ["Radio City", "India", "private FM radio station", "3 July 2001", "Hindi",
     "English", "New Media", "May 2008", "PlanetRadiocity.com", "music portal",
     "news", "videos", "songs"]
}
\end{lstlisting}

\medskip
\hrulefill \par
\textbf{Input Format:}
\begin{lstlisting}[style=mypromptstyle, framerule=0pt]
Passage: ${passage}
\end{lstlisting}
\end{tcolorbox}
\caption{LLM prompt for Entity Extraction. This prompt aims for comprehensive entity identification beyond standard NER.}
\label{fig:prompt_entity_extraction_app_fig}
\end{figure*}

\begin{figure*}[htbp] 
\begin{tcolorbox}[
    colback=white, colframe=black, boxrule=0.5pt, sharp corners,
    fontupper=\small, 
    left=2mm, right=2mm, top=2mm, bottom=2mm,
    title=Proposition Extraction Prompt
    ]

    \textbf{Instruction:} \par 
    Your task is to analyze text passages and break them down into precise, atomic propositions using a specified list of named entities. A proposition is a fully contextualized statement that expresses a single unit of meaning with complete specificity about the relationships described. \par
    \textit{For each proposition:}
    \begin{enumerate} \itemsep0pt \parskip0pt 
        \item Extract a complete, standalone statement that preserves the full context
        \item Use ONLY the entities provided in the \texttt{named\_entities} list - do not introduce new entities
        \item Ensure each proposition contains only ONE claim or relationship
        \item Be extremely specific about which entities are involved in each relationship
        \item Maintain clear causal connections between related statements
    \end{enumerate}
    Respond with a JSON object containing a list of propositions, where each proposition is an object with: \par
    - \texttt{"text"}: The full proposition text as a complete, contextualized statement \par
    - \texttt{"entities"}: An array of entities from the \texttt{named\_entities} list that appear in that proposition \par
    \textit{Critical Guidelines:}
    \begin{itemize} \itemsep0pt \parskip0pt 
        \item ONLY use entities from the provided \texttt{named\_entities} list
        \item Make relationships explicit and specific - clarify exactly which entities relate to which other entities
        \item Clarify precisely which entity a modifier applies to (e.g., specify which product had "80\% improvement")
        \item Establish clear connections between related facts (e.g., "Adobe optimized their applications FOR THE M1 CHIP")
        \item Connect comparative statements to their specific reference points (e.g., "Adobe's applications on the M1 chip improved by 80\% compared to Intel-based Macs")
        \item Preserve temporal information and causal relationships between events
        \item Make each proposition stand alone with all necessary context
        \item Include ALL relevant entities from the \texttt{named\_entities} list in both the proposition text and entities array
        \item Ensure the collection of propositions captures ALL meaningful information in the passage
    \end{itemize}

    \hrulefill \par 
    \textbf{Demonstration:} \par
    \textit{Passage:} In 2020, after Apple launched the M1 chip, major software companies like Adobe optimized their applications, improving performance by up to 80\% compared to Intel-based Macs. \par
    \textit{Named entities:}
    \begin{lstlisting}[style=mypromptstyle, framerule=0pt] 
    ["Apple", "M1 chip", "2020", "Adobe", "Adobe's applications", "Intel-based Macs", "80% performance improvement"]
    \end{lstlisting}
    \par 
    \begin{lstlisting}[style=mypromptstyle, framerule=0pt] 
{
  "propositions": [
    {
      "text": "Apple launched the M1 chip in 2020.",
      "entities": ["Apple", "M1 chip", "2020"]
    },
    {
      "text": "Adobe optimized their applications specifically for the M1 chip after its launch.",
      "entities": ["Adobe", "Adobe's applications", "M1 chip"]
    },
    {
      "text": "Adobe's applications running on the M1 chip improved performance by up to 80% compared to the same applications running on Intel-based Macs.",
      "entities": ["Adobe", "Adobe's applications", "M1 chip", "80% performance improvement", "Intel-based Macs"]
    }
  ]
}
    \end{lstlisting}

    \hrulefill \par 
    
    \textbf{Input Format:} \par
    \begin{lstlisting}[style=mypromptstyle, framerule=0pt]
Passage: ${passage}
Named entities: ${entities_json_list}
    \end{lstlisting}
\end{tcolorbox}
\caption{LLM prompt for Proposition Extraction. This prompt emphasizes contextual completeness and adherence to pre-identified entities.}
\label{fig:prompt_proposition_extraction_app_fig}
\end{figure*}

\begin{table*}[t!] 
\centering
\caption{Statistics of Constructed Proposition Graphs per Dataset.}
\label{tab:graph_statistics_app}
\small 
\begin{tabular}{@{}lrrrrr@{}}
\toprule
Statistic & NQ & PopQA & MuSiQue & 2Wiki & HotpotQA \\
\midrule
\# Propositions & 55536 & 57624 & 59028 & 30099 & 53566 \\
\# Passage Nodes & 9633 & 8676 & 11656 & 6119 & 9811 \\
\# Entity Nodes & 62368 & 73577 & 76928 & 43444 & 75608 \\
\# Total Edges & 1.27M & 1.17M & 1.34M & 0.86M & 1.31M \\ 
\bottomrule
\end{tabular}
\end{table*}

\subsubsection{Entity Extraction Prompt}
\label{subsec:entity_prompt_app}
This prompt is designed for inclusive entity identification. Unlike strict Named Entity Recognition (NER) often used for triple extraction, this step aims to capture a broader set of concepts relevant for constructing rich propositions. It explicitly asks the LLM to identify named entities, dates, important generic entities, and entities involved in predicate relations. This provides a comprehensive list for the subsequent proposition generation phase, which only uses entities from this pre-identified set.
(The prompt is shown in Figure \ref{fig:prompt_entity_extraction_app_fig})

\subsubsection{Proposition Extraction Prompt}
\label{subsec:prop_prompt_app}
This prompt guides the LLM to decompose a passage into atomic, yet contextually complete, propositions. It strictly uses the entities identified in the previous step (Figure \ref{fig:prompt_entity_extraction_app_fig}). The core focus is on maintaining high fidelity by preserving complex relationships, conditions, and the full context, which are often lost or oversimplified in traditional triple extraction processes.
(The prompt is shown in Figure \ref{fig:prompt_proposition_extraction_app_fig})

\subsection{Proposition Graph Statistics}
\label{app:graph_stats}
The proposition graphs constructed for each dataset vary in size and complexity, reflecting the nature of the underlying corpora. Table \ref{tab:graph_statistics_app} provides key statistics for the graphs used in our experiments. These include the number of extracted propositions, the number of passage nodes (corresponding to unique passages in the corpus subset), the number of unique entity nodes identified, and the total number of edges in the constructed graph (encompassing entity clique edges, passage containment edges, and synonymy edges).

\subsection{Cost and Efficiency}
\label{app:cost_efficiency}
The offline indexing phase of PropRAG involves LLM-based proposition and entity extraction, as well as embedding computation. For embedding, we run a float16 version of NV-Embed-v2 on an NVIDIA RTX 4090 GPU. For proposition and entity extraction, we utilize the Llama-3.3-70B-Instruct model via Nebius AI Studio's API endpoint. Processing each passage for proposition and entity extraction takes approximately 2 seconds with this setup. As a concrete example, indexing the 11,656 passages from the MuSiQue dataset completed within approximately 40 minutes, at a monetary cost of around \$4 USD using the API.

The token cost for the offline LLM-based proposition extraction is an important consideration. Table \ref{tab:token_cost_comparison_app} compares the input and output token counts for PropRAG on the MuSiQue dataset against those reported for other structure-augmented RAG methods by \citet{gutierrez2025rag} for their respective offline knowledge structuring phases.

\begin{table}[htbp] 
\centering
\caption{Offline LLM Token Costs (Input/Output) for Knowledge Structuring on MuSiQue Dataset (Millions of Tokens). Baseline data from \citet{gutierrez2025rag}.}
\label{tab:token_cost_comparison_app}
\small
\begin{tabular}{@{}lcc@{}}
\toprule
Method & Input Tokens (M) & Output Tokens (M) \\
\midrule
RAPTOR & 1.7 & 0.2 \\
HippoRAG 2 & 9.2 & 3.0 \\
\textbf{PropRAG (Ours)} & \textbf{16.5} & \textbf{4.6} \\
LightRAG & 68.5 & 18.3 \\
GraphRAG & 115.5 & 36.1 \\
\bottomrule
\end{tabular}
\end{table}

PropRAG's token cost for proposition extraction is higher than methods like HippoRAG 2 (which uses OpenIE for triple extraction, often less LLM-intensive) or RAPTOR (which focuses on summarization). This is attributable to the detailed instructions and the generation of full-sentence propositions, which are richer but require more tokens. However, PropRAG's costs are considerably lower than methods like LightRAG and GraphRAG, which may involve more extensive LLM-based processing for their graph construction or summarization steps. The trade-off is between the upfront offline cost of generating high-fidelity propositions and the downstream benefits in retrieval accuracy and the avoidance of online LLM calls during retrieval. The online retrieval phase of PropRAG, involving PPR and beam search, is entirely LLM-free and computationally efficient, relying on pre-computed embeddings and graph operations.

\subsection{Entity Score Calculation from Paths}
\label{app:entity_score_calculation}

After the beam search identifies a set of high-scoring proposition paths (as detailed in Section \ref{sssec:beam_search}), PropRAG determines the importance of individual entities based on their participation in these paths. This entity scoring is crucial for generating the final set of seed nodes ($S_{final}$) used in the Stage 2 PPR (Section \ref{sec:methodology}). The scoring process adheres to the following principles:

\begin{enumerate}
    \item \textbf{Path Score Inheritance:} Each proposition within an identified path is considered to have the same relevance score as the overall path it belongs to.
    \item \textbf{Entity Score Aggregation:} An entity's total score is determined by summing the scores of all propositions (and thus, all paths) in which it appears. If an entity is part of multiple high-scoring paths or multiple propositions within a single path, its score accumulates, reflecting its centrality and repeated relevance.

    \item \textbf{Emphasis on Connecting Entities:} The scoring mechanism gives additional weight to entities that form crucial links within a reasoning path, particularly for synonymous connections.
    \begin{itemize}
        \item \textbf{Synonymous Connections Boost:} When a proposition $P_A$ (containing entity $E_A$) connects to proposition $P_B$ (containing entity $E_B$) via a synonymous link where $E_A \approx E_B$, the \textit{connected entity} ($E_B$ in $P_B$) receives an additional score increment equivalent to the path's score. This effectively elevates the importance of $E_B$, treating it as a strong continuation of a central concept from $P_A$. The rationale is that $E_B$ is vital for identifying the passage associated with $P_B$. The original connecting entity ($E_A$ in $P_A$) contributes its score through its presence in $P_A$ but does not receive this specific connection-based score enhancement itself. If $P_A$ was connected from a preceding proposition, its own central entities would have been accounted for similarly.
        \item \textbf{Exact Connections:} Entities that are shared exactly between two consecutive propositions in a path (forming an exact connection) naturally contribute to the score aggregation through their appearance in both propositions. Their role as direct bridges is thus inherently emphasized by the summation of scores from both propositions they are part of.
    \end{itemize}
    \item \textbf{Initial Proposition Entities:} For entities appearing in the very first proposition of a path (which do not have a preceding "connection" within that path), their initial relevance is captured through the "exploration seeds" ($S_{initial}$). Many entities from these initial top query-relevant propositions are directly considered as exploration seeds. This ensures their potential importance is factored into the final seed set, even if they don't benefit from the connection-based score enhancements that apply to entities deeper within a path.
\end{enumerate}

Following the aggregation of scores for all entities involved in the discovered paths, the entities are ranked by their total accumulated scores. This ranked list is then used to select the top-$B_{beam}$ "exploitation seeds." These exploitation seeds, rich in path-derived relevance, are combined with the "exploration seeds" ($S_{initial}$) to form the final seed set $S_{final}$ for the concluding PPR stage, ensuring a comprehensive and robust final ranking of evidence passages.
\end{document}